\documentclass{article}
\usepackage{spconf,times,amsmath,graphicx,amssymb}

\usepackage{tikz}
\usetikzlibrary{arrows,automata}

\title{ Vision-based Engagement Detection in Virtual Reality }
\name{Ghassem Tofighi,  Kaamraan Raahemifar, Maria Frank, Haisong Gu,\thanks{This research is patented by Konica Minolta Laboratory U.S.A.,Inc.,
2855 Campus Dr \#100, San Mateo, CA 94403, USA }}
\address{Konica Minolta Laboratory U.S.A., Inc., San Mateo, CA, U.S.A., \\
	 Ryerosn University, Toronto, ON, Canada \\ 
	 Stanford University, Stanford, CA, U.S.A \\
\tt\small	gtofighi@ryerson.ca, kraahemi@ee.ryerson.ca, mfrank0@stanford.edu, haisong.gu@hl.konicaminolta.us }

\begin{document}
%
\maketitle
\begin{abstract}
User engagement modeling for manipulating actions in vision-based interfaces is one of the most important case studies of user mental state detection. In a Virtual Reality environment that employs camera sensors to recognize human activities, we have to know when user intends to perform an action and when not. Without a proper algorithm for recognizing engagement status, any kind of activities could be interpreted as manipulating actions, called "Midas Touch" problem. Baseline approach for solving this problem is activating gesture recognition system using some focus gestures such as waiving or raising hand. However, a desirable natural user interface should be able to understand user's mental status automatically. 

In this paper, a novel multi-modal model for engagement detection, DAIA \footnote{Disengagement, Attention, Intention, Action}, is presented. using DAIA, the spectrum of mental status for performing an action is quantized in a finite number of engagement states. For this purpose, a Finite State Transducer (FST) is designed. This engagement framework shows how to integrate multi-modal information from user biometric data streams such as 2D and 3D imaging.
FST is employed to make the state transition smoothly using combination of several boolean expressions. Our FST true detection rate is 92.3\% in total for four different states. Results also show FST can segment user hand gestures more robustly.
\end{abstract}
\section{Introduction}
User mental status detection has variety of applications in human activity recognition systems. Without a proper algorithm for detecting human intention to interact, the vision based system is always on, therefore any kind of activity may interpreted as an interaction \cite{kjeldsen2003dynamically}.
Group meetings which are frequent business events is modeled as a case study. In this case study, among all available data streams, a combination of tracking 3D skeleton data is combined for user engagement detection in meetings.  Multiple binary classifiers are implemented to detect user intention for performing an action. The output of these binary classifiers are used to create transition and guard conditions in FST. 
Characteristics of engagement will be discussed and biometric data which can be used for this purpose will be introduced. 3D skeleton tracking will be introduced as one of the channels of biometric information for engagement detection. Although we just use this only channel of biometric data, experiment results show we still can predict engagement with high accuracy. 

In addition, DAIA, the FST of engagement detection helps system to flow among states smoothly. FST is a predefined structure based on our knowledge of human activities that helps system predict engagement state more accurately. Furthermore, FST algorithm is computational efficient. This property of FST allows achievable on-line and real-time performance.

\subsection{Related Works}
Engagement has been investigated in various fields such as education, organizational behavior, work, or media. Engagement is defined as the value that a participant in an interaction attributes to the goal of being together with other participant(s) and continuing interaction \cite{cowie2015cbar, salam2015multi}. It is also defined as the process by which two or more participants establish, maintain, and end their perceived connection directly related to attention \cite{salam2015multi,sidner2004look} ). “Effort without distress” \cite{stajduhar2010patient}, “A meaningful involvement”\cite{sidner2004look}, Enabled through vigor, dedication, and absorption\cite{dael2012emotion}) are other definitions of engagement.

Body posture gives important information about engagement. Various approaches have been investigated based on body language analysis to improve human computer interaction. Intention to engage with an agent e.g. a robot \cite{vaufreydaz2015starting,klotz2011engagement}, or interactive display \cite{schwarz2014combining}, are some of these studies. Measuring the engagement intent is used in service robots to identify relevant gestures from irrelevant gestures which is known as Midas Touch Problem \cite{vaufreydaz2015starting,mead2011proxemic}. Another research interest is related to learner engagement with robotic companions or interfaces\cite{sanghvi2011automatic}.In addition, Intention to engage with a display for improving user identification is addressed in Schwarz et al. \cite{schwarz2014combining}. 

The role of body pose and motion in user’s interest detection using body tracking systems such as Kinect has been addressed in several research \cite{bianchi2013understanding, kleinsmith2013affective, bianchi2012can, asteriadis2009feature}.

A variety of studies strives for a multi-modal approach using some features of facial expression, body motion, voice, or seat pressure to elucidate on mental states. Benkaouar et al. \cite{benkaouar2012multi} is discussing gaze and upper body posture for engagement detetion.  Schwarz et. al \cite{schwarz2014combining} used combination of gaze, upper body and arm position for the purpose of intention detction in engagement. Vaufreydaz et. al \cite{vaufreydaz2015starting} used gaze and proxemics and Salam et. al \cite{salam2015multi} employed human state observation for engagement detection. Engell et al. used gaze and facial expression\cite{engell2007facial} and Balaban et al. \cite{balaban2004postural} employed weight, head, and upper body motion; Scherer et al. \cite{scherer2012generic} and Dael et. al \cite{dael2012emotion} discuss voice, face, posture. Using Finite State Machine (FSM) for multi-modal system modeling is addressed in multiple research\cite{johnston2002match},\cite{johnston2000finite},\cite{singletary2001learning},\cite{bourguet2003designing}. 

Mota used both neural networks for posture detection and Hidden Markov Models to detect engagement at an overall accuracy of 77\% which needs an expensive computation\cite{mota2003automated}. Michalowski et al. offered a spatial model combined with gaze tracking to detect user engagement with a robot receptionist \cite{michalowski2006spatial}.

All of the studies mentioned in this section discuss a very small set of potential classifiers. They also do not make full usage of the amount of qualitative research on non-verbal body language and its indication of mental states. In this research, we address these features for engagement modeling and detection.

\section{Engagement Modeling and Metrics}

Frank et. al\cite{frank2016engagement} have proposed a multi-modal engagement user state detection process. In our intended scenario, multiple people are within the operating range of the sensor, e.g. field of view of the camera. This module identifies the state of the participants on the engagement scale as disengaged up to involved action. 
Using the biometric information, the module identifies the person who exhibits a specific combination of classifiers of all categories. The classifiers, e.g. body posture components, are chosen based on research, relating them to attentiveness and engagement. 
The analysis is occurring on a frame-by-frame basis. Each frame is analyzed regarding all classifiers, e.g. if the person is 1) raising a hand, 2) facing a display, 3) leaning forward, 4) leaning backward, 5) uttering a feedback, 6)  slouching, or 7) changing position in the last 60 frames. Classifiers are evaluated as being exhibited or not exhibited in a specific frame as a binary value. 

Among all available biometric information such as gaze, voice and gesture to extend the proposed framework to detect engagement state of the user, we just use 3D joint information provided by a depth camera and NiTE SDK by Primesense. However, using more biometric channels of information will make system more accurate, our experiment results showed even use this only channel of information could result in a high performance user engagement detection system.

Upper body joints play important role in engagement detection. Multiple classifiers are designed to detect and classify upper body direction. In addition, hand movements such as raising hand, pointing, swiping, pushing or pulling are used to manipulate in vision-based interfaces. Therefore, different classifiers are designed to detect various hand movements such as raise hand above waist and also different levels of hand speed. These classifiers help to detect user intention for performing an action.

An action video with $T$ frames and $N$ joints in each frame can be represented as a set of 3D points sequence, written as $p = \{ x^{t}_{n}\in \mathbb{R}^{3} | n= 1,..,N,t=1,2,...,T \}$. The 3D sensor provides us fifteen joints,  and $T$ varies for different sequences. However, in our system $N = 10$, because our classifiers only use ten upper body joints from this set which are head, left and right shoulders, left and right elbows, left and right hand, torso and left and right hips. For each point, 3 dimensional position ${X,Y,D}$ is obtained.

The first basic step of feature extraction is to compute basic feature for each frame, which describes the pose information of every of these ten joints in a single frame. The second step is calculating of left and right hand speed information. This feature is obtained by calculating 3D distance that each hand moves in two consecutive frames.

The binary values for the individual binary classifiers are weighted based on the relative influence in the training data and summed up to . The engagement score thereby assumes a value between 0 and 1. 

\begin{figure}[h!]
	\centering
	\includegraphics[width=0.4\textwidth]{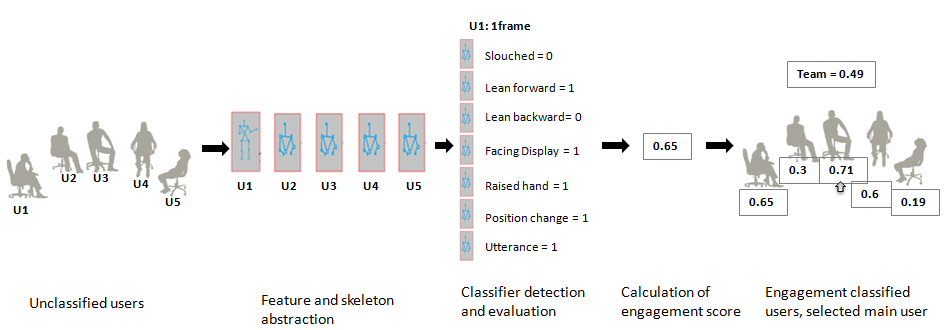}
	
	\caption{System Approach}
	\label{fig:system}
\end{figure}

Figure \ref{fig:system} shows how our system extract features from users and calculate engagement levels of each users.

The engagement score is calculated using $W'.G$ where $W$ is vector of weights $[w_{1}, w_{2}, w_{3}, ..., w_{n}]$, and $G$ is the vector of binary classifiers $[g_{1}, g_{2}, g_{3}, ..., g_{n}]$ such that $g_{1}, g_{2}, g_{3}, ..., g_{n}$ are 0 or 1 based on the output of the binary classifiers.

Table \ref{tab:BinaryClassifiers} gathers binary classifiers which are designed for this purpose. For each posture status, multiple binary classifiers are designed. The 0 or 1 output of these 37 classifiers are used to make our feature vector for intention to action classifier.

\begin{table}
	\scriptsize
	\begin{center}
		\begin{tabular}{|l|l|}
			\hline
			Posture Status & Binary Classifiers \\
			\hline\hline
			Hand Horizontal & Right of Body, Close to Body, Left of Body\\
			Hand Vertical &  Below Hip, Below Torso, Below Shoulder, Below Head\\
			Hand Depth & Back of Body, Close to Body, Front of Body\\
			Hand Speed & Stopped, Slow, Fast, Too Fast\\
			Body Direction & Facing Sensor\\
			Leaning & Lean back, No Lean, Lean Forward\\
			Special Postures & Hands folded, Hands on Head, Hands on Torso\\
			\hline
		\end{tabular}
	\end{center}
	\normalsize
	\caption{Binary classifiers for each posture status }
	\label{tab:BinaryClassifiers}
\end{table}

This classifiers are mostly designed based on heuristic information extracted from joints 3D location. For instance, body direction classifier is made using the normal vector of the plane containing right and left shoulders, and torso joint.

\section{DAIA: FST for Engagement Detection }
\begin{figure}[h!]
	\centering
	\includegraphics[width=0.35\textwidth]{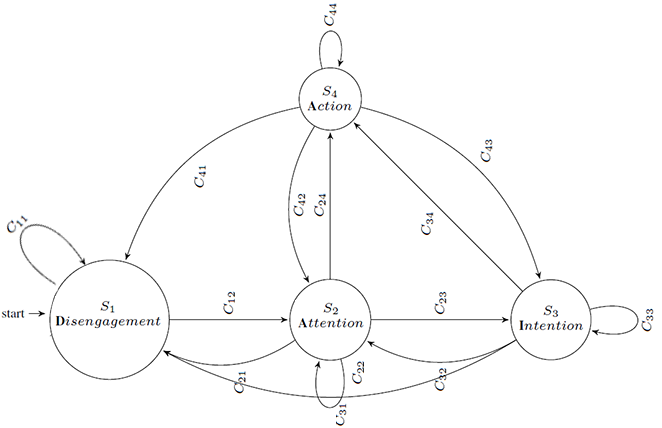}
	
	\caption{DAIA: FST for Engagement Detection}
	\label{fig:daia-framework}
\end{figure}

DAIA is a frame-based engagement detection system using an FST.  State machines are the description of a life cycle of a system. They can describe the different states of the lifeline, the events which influence it, and what it does when a particular event is detected at any states as the transition condition for particular state change. They offer the complete specification of the dynamic behavior of the system. Figure \ref{fig:daia-framework} shows the outline of this framework.

In order to increase efficiency and accuracy of our algorithm we implemented a Finite State Transducer (FST). A state is a description of the mental state or engagement of the user that is anticipated to change over time. A transition is initialized by a change in condition that results in a change of state. For example, when using a gesture recognition system to find out meaningful gestures of users, swiping or pointing can happen in some states such as in the \textit{Action} state and similar gestures in the \textit{Attention} state will be ignored or interpreted differently. In this research, we model engagement states as a  finite state transducer with four different states: \textbf{Disengagement}: User is disengaged from screen or the target person/object. \textbf{Attention}: User has attention, but doesn't have intention to do any actions. \textbf{Intention}: User has intention to do some action, but still not doing it. \textbf{Action}: User is performing an action.

A finite state transducer is a sextuple $(\Sigma, \Gamma, S, s_0, \delta, \omega)$, where: $\Sigma$ is the input alphabet (a finite non-empty set of symbols). $\Gamma$ is the output alphabet (a finite, non-empty set of symbols). $S$ is a finite, non-empty set of states.$s_0$ is the initial state, an element of $S$. $\delta$ is the state-transition function: $\delta: S \times \Sigma \rightarrow S$. $\omega$ is the output function. If the output function is a function of a state and input alphabet ($\omega: S \times \Sigma \rightarrow \Gamma$) that definition corresponds to the \textit{Mealy model}, and can be modeled as a \textit{Mealy machine}. 

Some hypotheses are considered in designing this FST: 
\textit{Engagement states change in a specific order}: This property describes the FST design. It starts with disengagement (Initial State).  All states can be transited to disengagement, but there is always a chain of ordered state transition for other states of engagement. 

\textit{We may modify the detected state based on conditions of FST}: States should transit smoothly. FST smooths state transition that helps better gesture segmentation. We don't change state just based on Intention to act or disengagement classifier. We know the human activities are continues, so, using some protection which is called guard conditions we smoothly change states. Furthermore, engagement state classifier is memoryless. it may report wrong engagement state based on the current biometric properties of the states. However, FST keeps record of engagement states and help relabeling the frames more accurately. 		

Table \ref{tab:transitionTable} describes \textit{Transition Conditions}, $C_{nm}$, for these state changes. $C_{nm}$ is the combination of event triggering the transition, the target state, guard and actions as follows:

\begin{table}
	\begin{center}
		\begin{tabular}{|l|l|l|l|l|l|}
			\hline
			State& $S_{1}$ & $S_{2}$ & $S_{3}$ &  $S_{4}$   \\
			\hline\hline
			$S_{1}$ & $C_{11}$ & $C_{12}$ & - & -\\
			$S_{2}$ & $C_{21}$ & $C_{22}$ & $C_{23}$  & $C_{24}$ \\
			$S_{3}$ & $C_{31}$ & $C_{32}$ & $C_{33}$ & $C_{34}$  \\
			$S_{4}$ & $C_{41}$ &   $C_{42}$    & $C_{43}$ & $C_{44}$  \\
			\hline
		\end{tabular}
	\end{center}
	\normalsize
	\caption{Transition table}
	\label{tab:transitionTable}
\end{table}

$C_{11}$, $C_{21}$, $C_{31}$, $C_{41}$  : \textit{Body Direction} is not facing sensor or a \textit{Special Posture} such as \textit{Hand Folded} exists

$C_{22}$, $C_{33}$, $C_{44}$  : The output of binary classifiers and \textit{Intention to Act } classifier doesn't change.
	
$C_{12}$: \textit{Body Direction} is facing sensor. $C_{23}$: \textit{Intention to Act} classifier is triggered and output is 1, but both \textit{Hand Speed} classifiers are stopped in at least one frame of each window of predefined number of consecutive frames. This window frame is a guard to protect state from transition because of small movements of hand which are not actions.  $C_{32}$: \textit{Intention to Act} classifier converts from 1 to 0 for more than predefined consecutive frames. This window of frames protects transition from action to Intention for small position changes that make \textit{Intention to Act} classifier zero. $C_{34}$: \textit{Intention to Act} classifier is triggered and output is 1, also at least one of \textit{Hand Speed} classifiers for detecting \textit{slow} or \textit{fast} movements is 1 for a predefined number of consecutive frames. $C_{43}$:  \textit{Intention to Act} classifier is triggered and output is 1, but both of \textit{Hand Speed} classifiers for detecting \textit{stopped} is 1 for a predefined number of consecutive frames.
After each transition, if FST recognizes the labels that is assigned to some frames are wrong, it can change reconsider and modify them by relabling. In addition, based on analysis of the speed signal of the hand, FST will relabel the frames when user starts moving hand to raise his hand. That helps we have correct segment of gesture for our gesture recognition system. 
\section{Experiment Results}
DAIA framework was implemented in C++ on Windows workstation. We used ASUS Xtion Pro to capture depth images and track skeletons using Primesense NiTE SDK. The system ran at 30 frames per second.
As it is mentioned in Table \ref{tab:BinaryClassifiers} we created 37 binary classifiers. For each posture status, multiple binary classifiers are designed. The 0 or 1 output of these 37 classifiers is used to make our feature vector, $G$, for intention to action classifier.  Furthermore, we need to define $W$ or vector of weights to calculate engagement score and afterwards we should define a threshold to classify the frame as intention to act or disengagement similar to the procedure proposed in \cite{schwarz2014combining}. It needs extensive research on different body postures to calculate these weights. Furthermore, putting constant weight values for different classifiers may result in wrong classification for complex body postures. Therefore, instead of defining constant values for the weight vector, we used SVM\cite{chang2011libsvm} with linear kernel for training our intention to act classification. We used $G$ as the feature vector for training and testing our SVM.

To train the classifier, a simple "Catch the Box!" game is designed. In this game, we used hand tracking algorithm implemented in NiTE middle-ware by Primesense to move the cursor on screen. A solid rectangle randomly appears on the screen and user should move the cursor on rectangle area to receive points. The game has 3 stages which are "Getting Ready", "Play" and "Stop". The binary classifier outputs of table \ref{tab:BinaryClassifiers} in "Play" mode of the game are combined as a series of  0 and 1 and used as "Intention to Act" feature vectors to train an binary SVM with linear kernel. The output of classifiers in "Getting Ready" and "Stop!" stages of game creates feature vectors of the SVM when "Intention to Act" is not present. 

We captured and labeled 23,210 frames from 5 different subjects that played the game separately. 5,000 frames are used for training and the remaining used for testing the classifier. This classifier performance was 86.38\%. The frame is classified as \textit{Intention to Act} or not.  
FST helps to relabel the frame based on the current state properties and guards. In our experiment, we asked 30 different users to hear random order of commands from a list of actions such as "raising hand" or "swiping right to left" and perform them in front of a screen.

\begin{figure}[h!]
	\centering
	\includegraphics[width=0.45\textwidth]{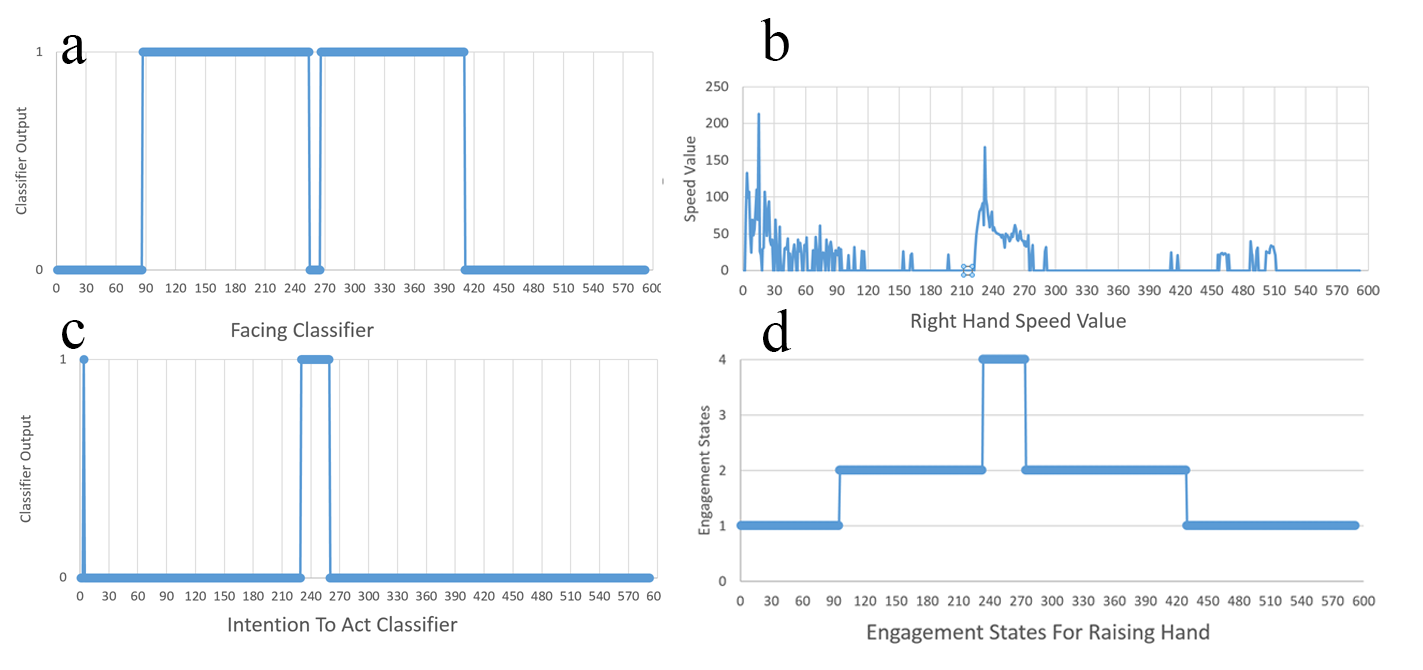}
	\caption{Engagement state detection using FST for raising and putting down right hand in 600 frames: a) Facing classifier b) Right hand speed value c) Intention to Act Classifier d) Engagement states for raising and putting down right hand}
	\label{fig:result}
\end{figure}

Afterwards, each recorded frame is labeled in four different engagement states from Disengagement to Action by an expert and used as ground truth. Our FST performance is calculated based on the number of correct states reported by FST after relabeling the frames and also ground truth labeled manually. In total, 165,422 frames are labeled to each engagement states. The results are gathered in table \ref{tab:FST-Results}. Figure \ref{fig:result} shows engagement state detection using FST and combinations of boolean operations on raising and putting down right hand.

\begin{table}
	\scriptsize
	\begin{center}
		\begin{tabular}{|l|l|}
			\hline
			State & FST Performance \\
			\hline\hline
			Disengagement & 97.3\%\\
			Attention &   87.2\%\\
			Intention & 90.8\%\\
			Action & 94.2\%\\
			\hline
			\textbf{Total} &  \textbf{92.3\%} \\
			\hline
		\end{tabular}
	\end{center}
	\normalsize
	\caption{Performance of  FST}
	\label{tab:FST-Results}
\end{table}

\section{Discussions and Conlutions}
In this paper, a novel multi-modal model for engagement is introduced. Based on this model, a combination of tracking 3D gesture data is employed for user engagement detection. Therefore, the spectrum of mental status for performing an action is quantized in a finite number of engagement states. Furthermore, a finite state transducer (FST) with the following engagement states is proposed: Disengagement, Attention, Intention, Action. Results show our \textit{Intention to Act} performance is 86.3\%. In addition, FST relables some of those labels based on the history of engagement states and guard conditions. The performance of our FST for labeling the frames correctly is 92.3\%.The processing time for each frame is less than 10ms which indicates real-time usability of our algorithm.

In future research, we expect using other channels of biometric information such as voice and facial data such as gaze. We may reach even higher true detection rates using extra channels of information. In addition, by using multi-camera and calculating engagement state for each of the audience in a  meeting room we will be able to detect the main operator and give the control of the vision based interface to that participant.


\end{document}